\documentclass[journal]{IEEEtran}

\usepackage{graphicx}
\usepackage{dcolumn}
\usepackage{bm}
\usepackage{algorithm}
\usepackage{algpseudocode}
\usepackage{algorithmicx}
\usepackage{booktabs}
\usepackage{rotating}
\usepackage[table,xcdraw]{xcolor}
\usepackage{cite}

\begin{document}
\title{Deep neural network initialization with decision trees}
\author{
    \IEEEauthorblockN{K. D. Humbird\IEEEauthorrefmark{1}\IEEEauthorrefmark{2}, J. L. Peterson\IEEEauthorrefmark{1}, R. G. McClarren\IEEEauthorrefmark{3}} \\
    \IEEEauthorblockA{\IEEEauthorrefmark{1}Lawrence Livermore National Laboratory, 7000 East Ave, Livermore, CA 94550} \\
    \IEEEauthorblockA{\IEEEauthorrefmark{2}Department of Nuclear Engineering, Texas A \& M University, 3133 TAMU, College Station, TX 77843}
\IEEEauthorblockA{\IEEEauthorrefmark{3}Department of Aerospace and Mechanical Engineering, University of Notre Dame, 365 Fitzpatrick Hall, Notre Dame, IN 46556}
}

\maketitle

\begin{abstract}
In this work a novel, automated process for constructing and initializing deep feed-forward neural networks based on decision trees is presented. The proposed algorithm maps a collection of decision trees trained on the data into a collection of initialized neural networks, with the structures of the networks determined by the structures of the trees. The tree-informed initialization acts as a warm-start to the neural network training process, resulting in efficiently trained, accurate networks. These models, referred to as ``deep jointly-informed neural networks'' (DJINN), demonstrate high predictive performance for a variety of regression and classification datasets, and display comparable performance to Bayesian hyper-parameter optimization at a lower computational cost. By combining the user-friendly features of decision tree models with the flexibility and scalability of deep neural networks, DJINN is an attractive algorithm for training predictive models on a wide range of complex datasets.

\end{abstract}

\begin{IEEEkeywords}
Bayes methods, decision trees, multilayer neural networks, neural networks
\end{IEEEkeywords}

\IEEEpeerreviewmaketitle

\section{Introduction}

\IEEEPARstart{D}{eep} neural networks are quickly becoming one of the most popular tools in machine learning due to their success at solving a wide range of problems-- from language translation \cite{google_language,signlanguage}, to image recognition \cite{imagerec_1,imagegen,objectrec}, to playing Atari \cite{google_atari,BayesianDropout_Gal}. Neural networks trained via supervised learning are capable of discovering subtle relationships between variables that make them well-suited for creating ``surrogate'' models for complex physical systems. Surrogate models approximate complicated response surfaces by interpolating between a set of sparse data that is typically expensive to acquire. The models provide a method for studying a continuum of designs rapidly, without resorting to costly computer simulations or experiments. Many machine learning algorithms can be used to create surrogates, but neural networks offer several distinct advantages: they are scalable to large volumes of high dimensional data, have low memory demands, and can be readily updated as new data becomes available. 

Despite the flexibility of neural networks, the application of deep learning to studying physics-based problems has been slow to increase in popularity. In part, the limited use of neural networks by non-experts is due to the difficulty of training an accurate model. There are an infinite number of design options, including the activation function, learning rate, regularization methods, and the network architecture: the number of hidden layers and the number of neurons in each layer. Often, changes in these settings can yield wildly different results. Datasets of interest for physics-based systems are often from high dimensional design spaces that are under-sampled and represent complex, nonlinear processes. The choice of neural network architecture for such data can significantly impact the training efficiency and accuracy of the model, and there exist few guidelines for determining appropriate settings that are robust across a multitude of problems. 

In many cases, simple machine learning algorithms can produce reasonably accurate surrogate models with minimal effort from the user. For example, decision tree-based algorithms, such as random forests or extremely randomized trees, have been successful at modeling a variety of physics-based datasets \cite{Breiman,Chen,extratrees}. Tree-based models are robustly accurate and have few hyper-parameters that need to be tuned, making them convenient ``black box'' algorithms. However, traditional trees are confined to on-axis splits, limiting the accuracy of the model, and the memory demands for storing an ensemble of trees is high for complex data. 

To create a black box neural network, the user-friendly features of tree-based models can be combined with the accuracy, flexibility, and scalability of deep neural networks. 
Several studies have explored the possibility of mapping decision trees and random forests to neural networks \cite{2hiddenlayer,oldNNweightinitial,neuraldecisiontree,neuralRF,neuralrf2}. One particularly successful approach maps trees to equivalent two hidden layer neural networks, with the number of neurons in each layer related to the number of leaves in the decision tree \cite{2hiddenlayer,neuralRF}. The mapping ``warm starts'' the neural network training process by initializing the network in a state that performs similarly to the decision tree; after additional training, the neural networks achieve higher accuracy than the original tree-based model. Although the two hidden layer models perform well for moderately-sized datasets, the networks can become quite wide for high-dimensional nonlinear regression problems with complex decision trees, making subsequent training difficult for limited-size datasets. 

While it is possible to fit any function with a sufficiently wide, shallow neural network \cite{univapprox}, studies suggest that deep networks often perform better than wide networks with a similar number of neurons \cite{GoodfellowML}. Including more hidden layers allows for higher levels of interaction between parameters, thus deep networks can discover nonlinear relationships not discernible with only two hidden layers. Based on this observation, we propose a novel mapping from decision trees to deep neural networks. The mapping produces a network with a specific number of hidden layers, neurons per hidden layer, and a set of initial weights that reflect the decision tree structure. The neural network is subsequently trained using back-propagation to optimize predictive performance. The algorithm is called ``deep jointly-informed neural networks'', or DJINN, as the final neural network is informed by an underlying decision tree model and the standard training method of back-propagation.

In the following sections, DJINN is described in detail and compared to a variety of other neural network models for regression and classification datasets. In section II, the algorithm for mapping from trees to initialized neural networks is presented and illustrated with a few examples. In section III, DJINN is presented as a ``warm start'' method for training deep neural networks and is compared to other warm start and weight initialization techniques. 
Section IV compares DJINN, which determines the neural network architecture based on the structure of a decision tree, to a Bayesian hyper-parameter optimization method for selecting an appropriate architecture. Although DJINN does not attempt to optimize the architecture of the neural network, it displays comparable performance to optimized architectures at a significantly lower computational cost. Overall, DJINN is observed to be a robustly accurate and user-friendly method for creating deep neural networks to solve a variety of classification and regression tasks. 

\section{Deep Jointly-Informed Neural Networks}
The DJINN algorithm determines an appropriate deep neural network architecture and weight initialization that utilizes the dependency structure of a decision tree trained on the data. The algorithm can be broken into 3 steps: constructing the ensemble of decision trees, mapping from trees to neural networks, and fine-tuning the neural networks via back-propagation. In the following sections, each step is presented in detail and the mapping is illustrated with a few simple examples.

\subsection{Decision tree construction}
The first step of the DJINN algorithm is the construction of the decision tree-based model. This can be a single decision tree that will result in a single neural network, or an ensemble of trees, such as random forests \cite{Breiman}, that will produce an ensemble of neural networks. The depth of the trees is often limited to avoid the creation of excessively large neural networks; the maximum tree depth is a hyper-parameter that should be tuned for each dataset.

\subsection{Mapping decision trees to deep neural networks}
The DJINN algorithm chooses a deep neural network architecture and a set of initial weights based on the structure of a decision tree. The mapping is not intended to reproduce the decision tree, but instead takes the decision paths as guidance for the network architecture and weight initialization.

While neural networks are initialized layer by layer, decision trees are typically stored by decision path. The paths begin at the top branch of the tree, and follow the left, and then the right, side of every decision until a leaf (prediction) is reached. The manner in which trees are stored makes them difficult to navigate according to depth, but simple to traverse recursively. When mapping from tree to neural network, it is easiest if the structure of the tree is known before initializing neural network weights, thus the decision paths are recursed through twice: first to determine the structure, then to initialize the weights. 

The primary branch of the tree is defined as the $l=0$ level. The levels then increase from $l=[1,D_t]$ where $D_t$ is the maximum tree depth, often specified by the user. The maximum branch depth is defined as $D_b=D_t-1$, as the last level of a decision tree contains only leaves. The mapped neural network has $D_t$ total layers: an input layer at $l=0$, $D_b$ hidden layers, and an output layer. The output layer contains one neuron per label for multi-label classification, or one neuron for single-output regression problems. 
Multi-output regression is accommodated by performing the mapping on multi-output decision trees \cite{sklearn}, and including one neuron per target variable in the output layer.

Algorithm \ref{djinn} outlines the process of initializing the DJINN network for a single tree. If an ensemble method is desired, a random forest or extremely randomized tree model can be used, and the mapping is repeated for each tree to create an ensemble of neural networks. 

\begin{algorithm}
\begin{algorithmic}[1]
\caption{DJINN Tree to Neural Network Mapping}\label{djinn}
\State Recurse through paths of the decision tree:
\Statex\hspace{\algorithmicindent} $\cdot$ Determine max branch depth ($D_b$)
\Statex\hspace{\algorithmicindent} $\cdot$ Count number of branches at each level $N_{b}(l)$ 
\Statex\hspace{\algorithmicindent} $\cdot$ Record max depth each input occurs as a branch: \Statex\hspace{\algorithmicindent} $\,$ $L^{\mathrm{max}}_i$
\vspace{2mm}
\Statex $\triangleright$ For a max branch depth $D_b$, there will be $D_b$ hidden layers, an input layer with N$_{\mathrm{in}}$ neurons, and an output layer with N$_{\mathrm{out}}$ (regression) or N$_{\mathrm{class}}$ (classification) neurons in the neural network. Each hidden layer will have $n(l)$ neurons, where
\begin{equation}
n(l)=n(l-1)+N_{b}(l)
\label{eq:n}
\end{equation}
This ``copies'' the previous hidden layer and adds ``new'' neurons for each branch in the current level of the tree.
\vspace{2mm}

\State Create arrays $W^l$ of dimension $n(l)$ x $n(l-1)$, $l$=1,...,$D_b$, and $W^{D_{b+1}}$ with dimension $n(D_b)$ x $N_{\mathrm{out}}$ (or $N_{\mathrm{class}}$) to store initial weights. Initialize arrays to 0.
\State For each input $i$=0,1,...,$N_{\mathrm{in}}$-1:
\Statex\hspace{\algorithmicindent} $\cdot$ Set $W_{i,i}^l$ = 1 for $l <$ $L^{\mathrm{max}}_i$
\Statex $\triangleright$ This ensures input values are passed through hidden layers until the decision tree no longer splits on them.

\State Recurse through decision paths of the tree:
\Statex\hspace{\algorithmicindent} \textbf{For} levels $l$=1,...,$D_b$:
\Statex\hspace{\algorithmicindent}\hspace{\algorithmicindent} \textbf{For} each node $c$ in level $l$:
\Statex\hspace{\algorithmicindent}\hspace{\algorithmicindent}\hspace{\algorithmicindent} $\cdot$Define $p$ as the neuron created by the
\Statex\hspace{\algorithmicindent}\hspace{\algorithmicindent}\hspace{\algorithmicindent} parent branch
\Statex\hspace{\algorithmicindent}\hspace{\algorithmicindent}\hspace{\algorithmicindent} \textbf{If} $c$ = branch:
\Statex\hspace{\algorithmicindent}\hspace{\algorithmicindent}\hspace{\algorithmicindent}\hspace{\algorithmicindent}$\,$$\triangleright$ According to Eq. \ref{eq:n}, a new 
\Statex\hspace{\algorithmicindent}\hspace{\algorithmicindent}\hspace{\algorithmicindent}\hspace{\algorithmicindent}$\,\,$ neuron has been added to layer $l$

\Statex\hspace{\algorithmicindent}\hspace{\algorithmicindent}\hspace{\algorithmicindent}\hspace{\algorithmicindent} $\cdot$ Initialize $W_{\mathrm{new},p}^l$$\sim$ $\mathcal{N}(0,\sigma^2)$,
\Statex\hspace{\algorithmicindent}\hspace{\algorithmicindent}\hspace{\algorithmicindent}\hspace{\algorithmicindent} $\,$ connecting branch $p$ and new neuron
\Statex\hspace{\algorithmicindent}\hspace{\algorithmicindent}\hspace{\algorithmicindent}\hspace{\algorithmicindent} $\cdot$ Initialize $W_{\mathrm{new},c}^l$$\sim$ $\mathcal{N}(0,\sigma^2)$,
\Statex\hspace{\algorithmicindent}\hspace{\algorithmicindent}\hspace{\algorithmicindent}\hspace{\algorithmicindent} $\,$ connecting branch $c$ and new neuron

\Statex\hspace{\algorithmicindent}\hspace{\algorithmicindent}\hspace{\algorithmicindent} \textbf{If} $c$ = leaf:
\Statex\hspace{\algorithmicindent}\hspace{\algorithmicindent}\hspace{\algorithmicindent}\hspace{\algorithmicindent} $\cdot$ Initialize $W_{p,p}^l$ $\sim$ $\mathcal{N}(0,\sigma^2)$, $l$=$l$+1 ...$D_b$-1
\Statex\hspace{\algorithmicindent}\hspace{\algorithmicindent}\hspace{\algorithmicindent}\hspace{\algorithmicindent} $\cdot$ Initialize $W_{p,\mathrm{out}}^{D_b}$ $\sim$ $\mathcal{N}(0,\sigma^2)$

\Statex\hspace{\algorithmicindent}\hspace{\algorithmicindent}\hspace{\algorithmicindent}\hspace{\algorithmicindent} $\triangleright$ Classification: $\mathrm{out}$ = neuron for the class
\Statex\hspace{\algorithmicindent}\hspace{\algorithmicindent}\hspace{\algorithmicindent}\hspace{\algorithmicindent} $\triangleright$ Regression: $\mathrm{out}$ = output neurons
\end{algorithmic}
\end{algorithm}


The variance of the normal distribution used to initialize nonzero DJINN weights is $3$$/$$(n_{\mathrm{prev}}+n_{\mathrm{cur}})$, where $n_{\mathrm{prev}}$ and $n_{\mathrm{cur}}$ are the numbers of neurons in the previous and current hidden layers, respectively. Biases for each neuron are randomly sampled from the same distribution. This is a variant of the popular Xavier initializer \cite{Xavierweights,tflearn}. The variance of the distribution is designed to keep the scale of the gradients roughly the same in all layers of a deep neural network. Weights that are used to pass input variables or leaf values through the hidden layers are initialized to unity in order to preserve their value. 

\subsection{Optimizing the neural networks}
Once the trees have been mapped into initialized neural networks, subsequent tuning of the weights is carried out using back-propagation. For the examples presented in the following sections, the neural networks are trained using Google's deep learning software Tensorflow \cite{tensorflow2015}. The activation function used at each hidden layer is the rectified linear unit (ReLu), which generally performs well for deep neural networks \cite{relu,relu2} and can exactly retain the values of neurons in previous hidden layers. The Adam optimizer \cite{adam} is used to minimize the cost function, which is mean squared error (MSE) for regression, and cross-entropy with logits for classification \cite{crossentropy}.

\subsection{Examples mapping from trees to DJINN models}

\begin{figure}
\begin{center}
		\includegraphics[width=0.45\textwidth]{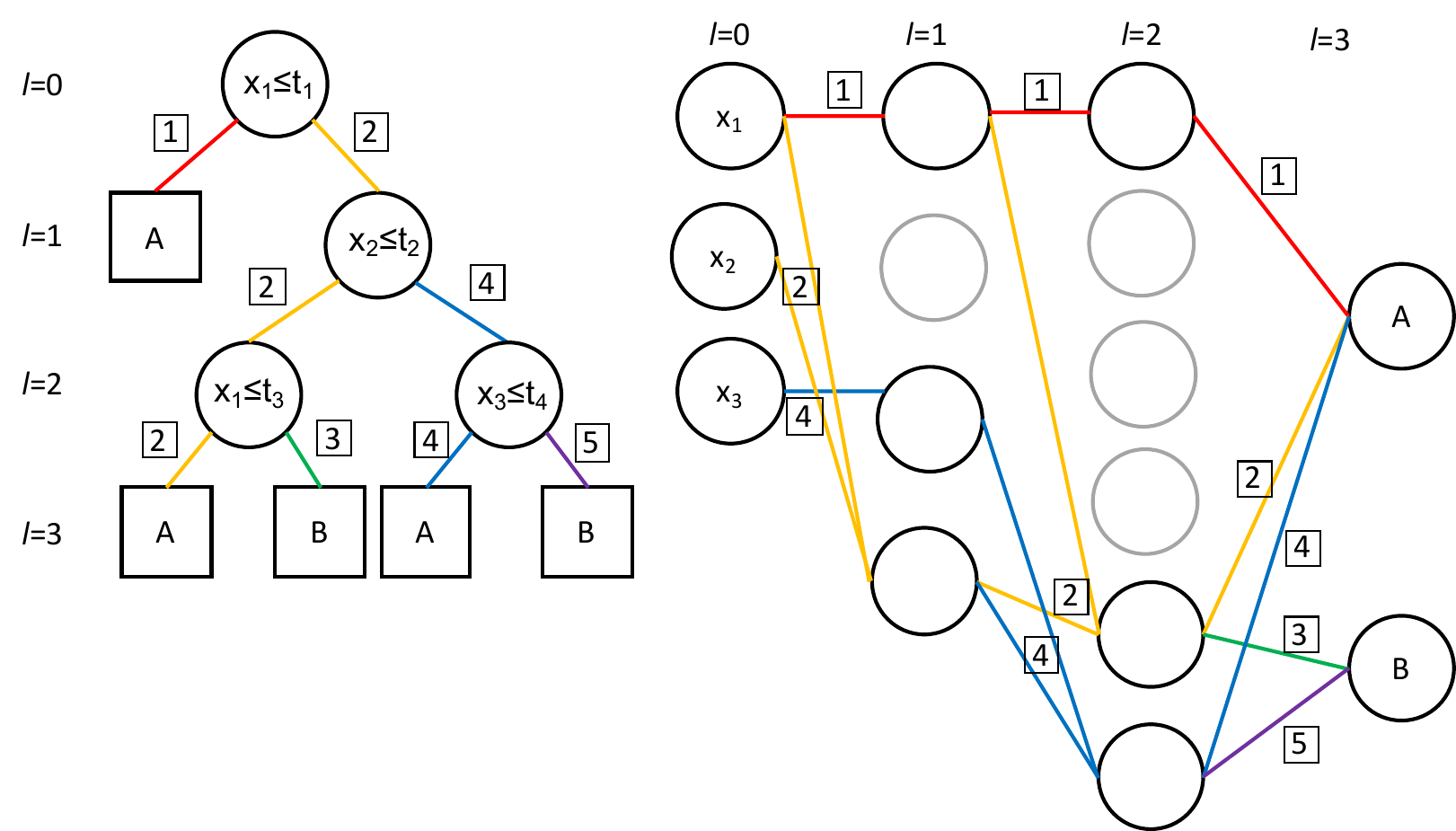}
		\caption{Illustration of the DJINN mapping for a simple decision tree. Gray neurons are initially unconnected; if biases are randomly initialized to negative values, the neurons cannot learn and are thus not included in the final DJINN architecture.}
		\label{fig:ttn1}
\end{center}
\end{figure}

Figure \ref{fig:ttn1} shows a simple example of a decision tree and the initialized DJINN neural network. Following the steps outlined in the algorithm, the mapping is performed as follows: 
\begin{enumerate}
\item \begin{itemize}
\item The maximum tree depth is $D_t=3$, as indicated by the numbers $l=0,...,3$. 
\item There are (1, 1, 2, 0) branches in each level of the tree
\item The maximum depth at which each input occurs as a branch is given by: $L^{\mathrm{max}}_i$ = (2, 1, 2) for x$_1$, x$_2$, and x$_3$.
\end{itemize}
\item The neural network architecture is shown in Fig. \ref{fig:ttn1}; initially all weights are set to zero. 

\item Set $W_{i,i}^l$ = 1 for $l <$ $L^{\mathrm{max}}_i$; this ``retains'' the input values through the hidden layers until they are no longer used as branches. In the figure, this step is represented by the horizontal red (labeled 1) and blue (2) connections for x$_1$ and x$_3$, respectively.

\item \begin{itemize}
\item Start with leftmost branch (red, 1). This node is a leaf, which uses x$_1$ to determine if the output is class A. The horizontal red connections propagate the value of the parent, x$_1$, through the hidden layers to the output layer, then connect to class A.

\item  Consider the yellow path (2) in the tree:
\begin{itemize}
\item For $l$=1 the node is a branch splitting on x$_2$, with parent x$_1$. One of the ``new'' neurons in $l$=1 of the neural network represents this decision. Connect x$_1$ and x$_2$ to this neuron (yellow, 2).
\item For $l$=2 there is a branch splitting on x$_1$; connect the parent (new neuron in $l$=1) and x$_1$ to a new neuron in $l$=2 (yellow, 2).
\item For $l$=3 there are two leaves, connect the parent (new neuron in $l$=2) to class A (yellow, 2) and B (green, 3). \end{itemize}

 \item Move to the rightmost path of the tree: 
 \begin{itemize}
 \item The $l$=1 layer, which created a new neuron that accepts x$_1$ and x$_2$ in $l$=1 of the network, has already been mapped.
\item For $l$=2 there is a branch splitting on x$_3$; connect the parent (new neuron in $l$=1 of the neural network) and x$_3$ to a new neuron in $l$=2 (blue, 4).
\item For $l$=3 there are two leaves; connect the parent (new neuron in $l$=2) to class A (blue, 4) or B (purple, 5).  \end{itemize}
\end{itemize}
\end{enumerate}

In step 4, all ``connections'' are non-zero weights initialized from the Xavier normal distribution as described previously, unless already initialized to unity. Qualitatively, the algorithm maps decision paths in the tree to decision paths through the network. Neurons which are not initially connected are randomly included in the final architecture; all biases are randomly initialized from a normal distribution, thus neurons with positive biases can be trained. The inclusion of extra degrees of freedom allows for the neural network to correct for inaccuracies in the decision tree during training. 


\begin{figure}
\begin{center}
		\includegraphics[width=0.4\textwidth]{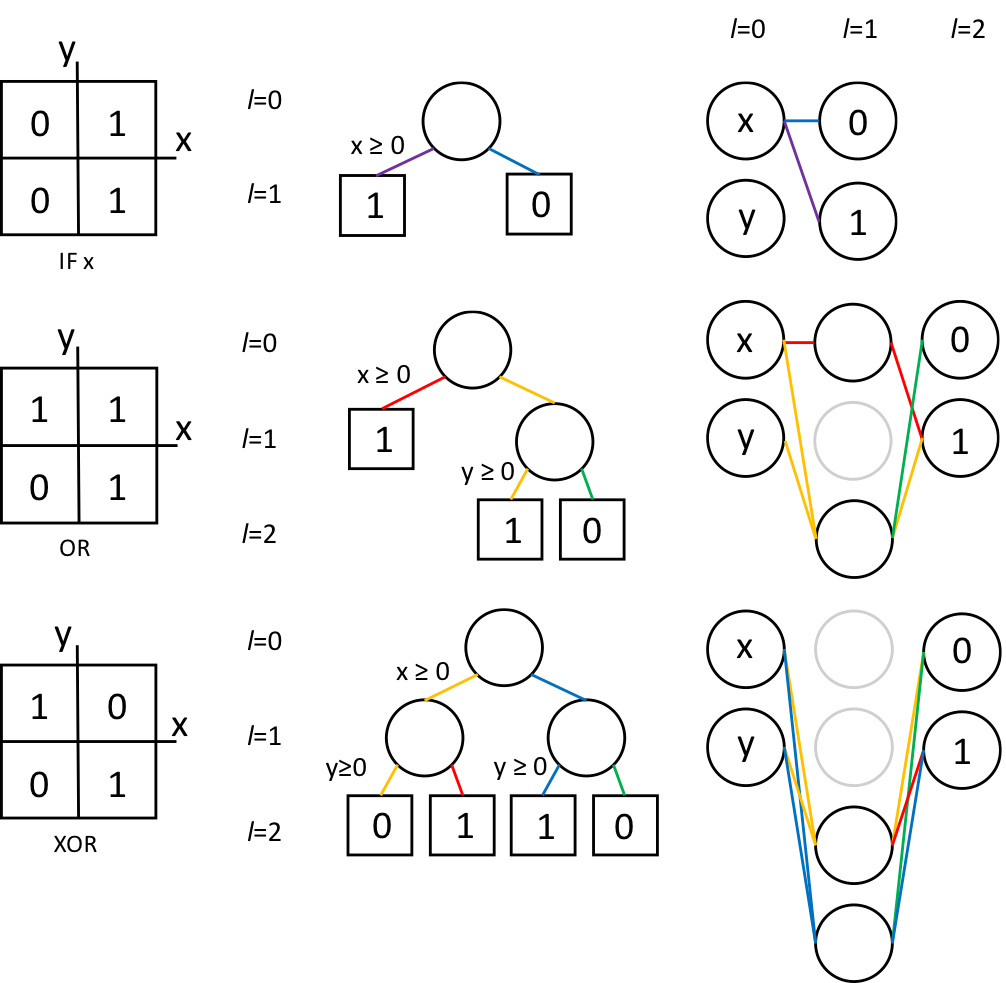}
		\caption{Truth tables, decision trees, and DJINN-initialized neural networks for logical operations IF(x), OR, and XOR. Decision paths in the tree are mapped to paths through the neural network, indicated by color. Gray neurons are initially unconnected; if biases are randomly initialized to negative values, the neurons cannot learn and are thus not included in the final DJINN architecture.}
		\label{fig:logic}
\end{center}
\end{figure}

As decision trees are sequences of logical operations, further insight into the mapping can be gained by considering how DJINN initializes networks to solve simple logic problems. Figure \ref{fig:logic} illustrates the decision tree and DJINN mapping for three logic operations that have unique decision tree structures: the IF, OR, and XOR statements. The connections in the initialized DJINN networks indicate nonzero weights and all biases are random. Gray neurons represent those that are not initially connected, but could be included in training if randomly assigned a positive bias.

For the “IF x” statement, the tree contains a single decision based on the value of x. DJINN reproduces this decision path by connecting the input x to either 0 or 1; knowing the value of x alone is enough to solve the problem.
Two decisions are needed to solve the “OR” problem: if x$\geq$0 the answer is 1, otherwise it needs to also consider the value of y. If x$<$0 and y$\geq$0 the answer is 1, otherwise the answer is 0. In the DJINN mapping of this tree, the value of x is passed directly to the output class 1, as shown by the red connections; mimicking the left side of the tree. To mimic the right side of the tree, both the values of x and y are passed to the last neuron in the hidden layer, which is then connected to classes 0 and 1. 
The XOR operation requires knowledge of x and y to determine the correct class. The DJINN initialization has two hidden neurons that receive both x and y, which then connect to the output layer. For the OR and XOR problems, the gray neurons that are randomly included can correct for errors in the decision tree. For simple logic operations, the presence of additional neurons is not necessary, but for complicated problems the decision tree is often too simple to accurately model the data. 

Since decision trees are a series of logical operations, DJINN initialization can also be viewed as such. When a branch splits into two additional branches, there is an XOR-like decision; when a branch splits into a branch and a leaf there is an OR-like decision, and when a branch splits into two leaves, there is an IF-like decision. This is illustrated in Fig. \ref{fig:ttn1}: the first decision is OR-like-- the red and yellow connections in the neural network from Fig. \ref{fig:ttn1} match those in Fig. \ref{fig:logic}. The next decision is XOR-like-- the yellow and blue connections between $l$=1 and $l$=2 in Fig. \ref{fig:ttn1} match those from the XOR in Fig. \ref{fig:logic}. Finally, there are two IF-like decisions, which connect the neurons from the final hidden layer to the outputs as shown by the blue/purple and yellow/green connections in Fig. \ref{fig:ttn1}.

Currently, the thresholds of the logical operations are tuned during training; a potential path for improving the algorithm is to encode the decision tree thresholds into the neural network initialization procedure.


\section{DJINN Performance}
The ease of use of the DJINN algorithm makes it an attractive method for general researchers to create neural network-based surrogate models for complex datasets. Unlike hyper-parameter optimization algorithms used to design neural networks \cite{genetic,reinforce}; DJINN does not require expensive searches through high-dimensional parameter spaces in order to determine a suitable neural network architecture and weight initialization. 

In the following sections, the performance of DJINN is compared to alternative methods for neural network design and initialization for a variety of regression and classification datasets. In section III A, the benefits of using DJINN as an ensemble method are explored, followed by a comparison to shallow neural networks initialized from decision trees in section III B. 
In section III C, the importance of the initial topology of the DJINN weights is illustrated by comparing DJINN to other initializations: densely connected topologies, and sparsely-connected initial weights that do not leverage the dependency structure of the data learned by a decision tree. The DJINN initialization is shown to provide a warm-start to the training process for a variety of datasets, allowing the models to achieve higher predictive performance than non-informative initialization techniques in a fixed amount of training time. 

\subsection{DJINN as an ensemble method}
The DJINN algorithm maps a decision tree to a deep neural network with an architecture and initial weights that reflect the dependency structure of the data learned by the tree. In practice, ensembles of decision trees, such as random forests \cite{Breiman} or extra-trees models \cite{extratrees} often exhibit significantly higher performance than individual decision trees. In the ensemble approach, each tree is trained on a random subset of the data and gains complementary knowledge about the relationship between the input and target variables. Each tree makes its own prediction for the target variables, and the model reports the mean prediction of the ensemble. Increasing the number of trees in the ensemble improves predictive performance up to some maximum number of trees, at which point the model begins to over-fit to the training data.

\begin{figure}
\begin{center}
		\includegraphics[width=0.5\textwidth]{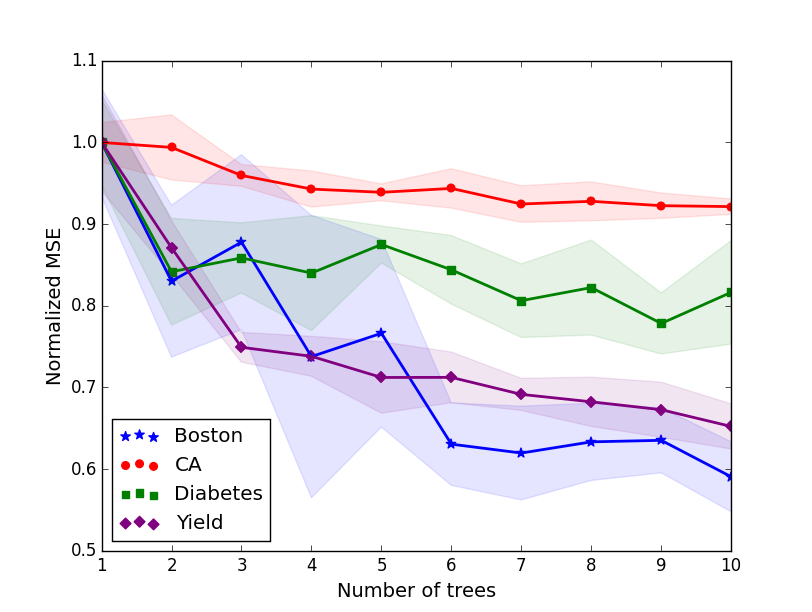}
		\caption{MSE (normalized to the MSE of one tree) as a function of the number of trees included in the DJINN ensemble for various regression datasets. The performance of the model improves as the number of trees is increased.}
		\label{fig:msevtree}
 \end{center}
\end{figure}

Similar to the random forest from which DJINN is mapped, the performance of DJINN improves as the number of trees included in the ensemble increases. Figure \ref{fig:msevtree} plots the predictive performance of DJINN as the number of tree-initialized neural networks increases; the mean squared error (MSE) is normalized by the MSE of the single-tree model. The bold line shows the mean value from a five-fold cross-validation score, and the error bars represent the standard deviation of the score. The cross-validation is performed by randomly splitting the data into training (80\%) and testing (20\%) groups with a fixed random seed, such that each model sees the five same permutations of training and testing data.

Included in Figure \ref{fig:msevtree} are three standard regression datasets: California housing prices \cite{cahouse}, Boston housing prices \cite{boston}, and diabetes disease progression \cite{diabetes}. DJINN is also tested on a novel database of inertial confinement nuclear fusion (ICF) implosion simulations \cite{langer}. The ICF dataset consists of 46,416 points Latin hypercube sampled from a nine-dimensional input space. The output of interest is the yield: the thermonuclear energy produced in the implosion. The yield response surface has proved challenging to fit with common machine learning algorithms \cite{Peterson2017}, as there are many nonlinear cliff- and peak-like features that are not well resolved by the data. A 300-tree random forest regressor \cite{Breiman} previously proved to be the most successful model, with a mean prediction error of approximately 10$\%$ and an explained variance score of 0.92. 

The models are trained with fixed hyper-parameter settings, summarized in Table \ref{table:hp}. Features in each dataset are scaled between (0,1) prior to training, but performance metrics such as MSE and mean absolute error (MAE) are reported in unscaled units, unless otherwise noted. 

For each of the regression datasets, the error of the DJINN model decreases with the number of trees included in the model; this behavior is characteristic of the random forest from which DJINN is mapped. In tree-based ensemble methods, there is typically a minimum number of trees that achieves low prediction error; adding more trees yields diminishing improvements in performance, and eventually leads to over-fitting. In the following sections, DJINN is always evaluated as an ensemble method with ten trees per model. 

Ensemble methods are becoming popular for various neural network applications; in particular, recent work has shown that attention models exhibit improved performance when treated as ensembles. Attention models are popular for exploiting dependencies between variables, particularly for time series and sequence data \cite{attention2,googleattention,attention1}. Rather than using a single attention model, it has been observed that using an ensemble of models, in which each model is initialized with a different structure to extract complementary information from the data, often leads to superior performance \cite{jayattention}. This is analogous to the improvements seen in DJINN, in which each neural network is initialized with a different dependency structure learned by the trees in the random forest. The importance of this dependency structure will be emphasized in the next sections.

\begin{table}[]
\centering
\caption{Neural network hyper-parameters used for each dataset.}
\label{table:hp}
\begin{tabular}{|c|c|c|c|c|}
\hline
\multicolumn{1}{|c|}{Dataset} & \multicolumn{1}{c|}{\# Epochs} & \multicolumn{1}{c|}{\begin{tabular}[c]{@{}c@{}}Learn. \\ Rate\end{tabular}} & \multicolumn{1}{c|}{\begin{tabular}[c]{@{}c@{}}Batch \\ Size\end{tabular}} & \multicolumn{1}{c|}{\begin{tabular}[c]{@{}c@{}}Max. Tree\\ Depth\end{tabular}}\\ \hline
Boston housing & 300 & 0.006 & 21  & 5\\
CA housing & 200 & 0.006 & 826 & 5 \\
Diabetes & 50 & 0.0001 & 1 & 5\\
ICF Yield & 300 & 0.008 & 1857 & 5  \\
Iris & 100 & 0.006 & 6  & 3\\
Digits & 300 & 0.003 & 72 & 3 \\
Wine & 50 & 0.004 & 8  & 3 \\
Breast cancer & 100 & 0.006 &  7 & 4 \\ \hline
\end{tabular}
\end{table}

\subsection{Comparison to shallow tree-initialized neural networks}
Other algorithms for mapping decision trees to two-hidden layer neural networks are observed to act as ``warm-starts'' to neural network training \cite{2hiddenlayer,neuralRF}. In the first two rows of Table \ref{table:reg}, DJINN is compared to two-hidden layer networks for the regression datasets presented in Figure \ref{fig:msevtree}. The first two rows of Table \ref{table:class} show the performance of the models for four standard classification datasets: the iris flower \cite{iris}, digits \cite{mnist}, wine \cite{wine}, and breast cancer \cite{breastcancer}. 
DJINN and the two-hidden layer model (abbreviated 2HL) are evaluated as ensemble methods; the trees are mapped from ten-tree random forests, and the ensemble prediction is the mean of the ten individual predictions. The networks are trained with the hyper-parameters summarized in Table \ref{table:hp} on five fixed permutations of training and testing data to produce cross-validation scores. The performance metrics include MSE, mean absolute error (MAE) and explained variance (EV) for regression, and recall, precision, and accuracy for classification. Student's t-tests between the MSE values for DJINN and the 2HL model give the p-value listed in the final column for each dataset.

\begin{table*}[]
\centering
\caption{Model performances for regression tasks. The mean and standard deviation of five-fold cross-validation metrics are reported for each model. The p-value is computed with a Student's t-test between the test MSE values for DJINN and the other models. Bold blue values highlight comparisons in which DJINN has a lower error than the other method with p$<$0.05; bold red values highlight when DJINN has higher error than the other method with p$<$0.05.}
\label{table:reg}
\begin{tabular}{lllllllll}
\multicolumn{1}{l|}{} & Boston &  &  & \multicolumn{1}{l|}{} & CA Housing &  &  &  \\ \hline
\multicolumn{1}{l|}{Model} & MSE & MAE & EV & \multicolumn{1}{l|}{p} & MSE & MAE & EV & p \\ \hline
\multicolumn{1}{l|}{DJINN} & 7.289$\pm$1.541 & 1.840$\pm$0.105 & 0.915$\pm$0.014 & \multicolumn{1}{l|}{} & 0.233$\pm$0.011 & 0.318$\pm$0.006 & 0.826$\pm$0.010 &  \\
\multicolumn{1}{l|}{2HL} & 8.393$\pm$4.568 & 1.965$\pm$0.346 & 0.903$\pm$0.059 & \multicolumn{1}{l|}{0.622} & 0.307$\pm$0.009 & 0.381$\pm$0.008 & 0.766$\pm$0.007 & {\color[HTML]{3531FF} \textbf{3.110E-06}} \\ \hline
\multicolumn{1}{l|}{Random-Dense} & 8.440$\pm$1.897 & 1.906$\pm$0.101 & 0.901$\pm$0.020 & \multicolumn{1}{l|}{0.323} & 0.247$\pm$0.011 & 0.327$\pm$0.007 & 0.816$\pm$0.009 & {\color[HTML]{3531FF} \textbf{0.004}} \\
\multicolumn{1}{l|}{Random-Sparse} & 7.326$\pm$0.707 & 1.898$\pm$0.062 & 0.914$\pm$0.009 & \multicolumn{1}{l|}{0.962} & 0.270$\pm$0.006 & 0.347$\pm$0.009 & 0.798$\pm$0.006 & {\color[HTML]{3531FF} \textbf{2.009E-4}} \\ \hline
\multicolumn{1}{l|}{Bayesian Opt.} & 7.556$\pm$0.815 & 2.034$\pm$0.068 & 0.910$\pm$0.007 & \multicolumn{1}{l|}{0.740} & 0.305$\pm$0.011 & 0.377$\pm$0.012 & 0.772$\pm$0.006 & {\color[HTML]{3531FF} \textbf{8.470E-06}} \\
 &  &  &  &  &  &  &  &  \\
\multicolumn{1}{l|}{} & Diabetes &  &  & \multicolumn{1}{l|}{} & Yield &  &  &  \\ \hline
\multicolumn{1}{l|}{Model} & MSE & MAE & EV & \multicolumn{1}{l|}{p} & MSE & MAE & EV & p \\ \hline
\multicolumn{1}{l|}{DJINN} & 3154$\pm$339.9 & 43.391$\pm$2.006 & 0.455$\pm$0.100 & \multicolumn{1}{l|}{} & 0.018$\pm$0.002 & 0.063$\pm$0.003 & 0.990$\pm$0.001 &  \\
\multicolumn{1}{l|}{2HL} & 3108$\pm$153.3 & 43.456$\pm$1.381 & 0.421$\pm$0.043 & \multicolumn{1}{l|}{{\color[HTML]{333333} 0.787}} & 0.031$\pm$0.005 & 0.088$\pm$0.012 & 0.983$\pm$0.003 & {\color[HTML]{3531FF} \textbf{8.380E-4}} \\ \hline
\multicolumn{1}{l|}{Random-Dense} & 3414$\pm$266.5 & 44.704$\pm$1.716 & 0.383$\pm$0.055 & \multicolumn{1}{l|}{0.215} & 0.021$\pm$0.001 & 0.067$\pm$0.003 & 0.989$\pm$0.001 & {\color[HTML]{3531FF} \textbf{0.045}} \\
\multicolumn{1}{l|}{Random-Sparse} & 3045$\pm$188.5 & 43.783$\pm$1.268 & 0.461$\pm$0.061 & \multicolumn{1}{l|}{0.547} & 0.049$\pm$0.007 & 0.111$\pm$0.011 & 0.973$\pm$0.003 & {\color[HTML]{3531FF} \textbf{9.880E-06}} \\ \hline
\multicolumn{1}{l|}{Bayesian Opt.} & 2376$\pm$107.1 & 38.895$\pm$1.519 & 0.584$\pm$0.044 & \multicolumn{1}{l|}{{\color[HTML]{FE0000} \textbf{0.001}}} & 0.023$\pm$0.003 & 0.081$\pm$0.008 & 0.988$\pm$0.001 & {\color[HTML]{3531FF} \textbf{0.020}}
\end{tabular}
\end{table*}

\begin{table*}[]
\centering
\caption{Model performances on classification tasks. The mean and standard deviation of five-fold cross-validation metrics are reported for each model. The p-value is computed with a Student's t-test between the test accuracy values for DJINN and the other models. Bold blue values highlight comparisons in which DJINN has a lower error than the other method with p$<$0.05; bold red values highlight when DJINN has higher error than the other method with p$<$0.05.}
\label{table:class}
\begin{tabular}{lllllllll}
\multicolumn{1}{l|}{} & Iris &  &  & \multicolumn{1}{l|}{} & Digits &  &  &  \\ \hline
\multicolumn{1}{l|}{Model} & Recall & Precision & Accuracy & \multicolumn{1}{l|}{p} & Recall & Precision & Accuracy & p \\ \hline
\multicolumn{1}{l|}{DJINN} & 0.987$\pm$0.020 & 0.980$\pm$0.029 & 0.983$\pm$0.025 & \multicolumn{1}{l|}{} & 0.973$\pm$0.010 & 0.977$\pm$0.008 & 0.976$\pm$0.009 &  \\
\multicolumn{1}{l|}{2HL} & 0.950$\pm$0.052 & 0.959$\pm$0.045 & 0.959$\pm$0.045 & \multicolumn{1}{l|}{0.144} & 0.971$\pm$0.015 & 0.971$\pm$0.015 & 0.972$\pm$0.015 & 0.549 \\ \hline
\multicolumn{1}{l|}{Random-Dense} & 0.982$\pm$0.019 & 0.975$\pm$0.027 & 0.978$\pm$0.023 & \multicolumn{1}{l|}{0.289} & 0.976$\pm$0.011 & 0.979$\pm$0.009 & 0.978$\pm$0.010 & 0.667 \\
\multicolumn{1}{l|}{Random-Sparse} & 0.988$\pm$0.011 & 0.979$\pm$0.019 & 0.983$\pm$0.015 & \multicolumn{1}{l|}{0.289} & 0.971$\pm$0.005 & 0.972$\pm$0.004 & 0.972$\pm$0.004 & 0.303 \\ \hline
\multicolumn{1}{l|}{Bayesian Opt.} & 0.980$\pm$0.015 & 0.980$\pm$0.014 & 0.978$\pm$0.016 & \multicolumn{1}{l|}{0.147} & 0.964$\pm$0.021 & 0.965$\pm$0.021 & 0.965$\pm$0.020 & 0.240 \\
 &  &  &  &  &  &  &  &  \\
\multicolumn{1}{l|}{} & Breast Cancer &  &  & \multicolumn{1}{l|}{} & Wine &  &  &  \\ \hline
\multicolumn{1}{l|}{Model} & Recall & Precision & Accuracy & \multicolumn{1}{l|}{p} & Recall & Precision & Accuracy & p \\ \hline
\multicolumn{1}{l|}{DJINN} & 0.960$\pm$0.012 & 0.954$\pm$0.019 & 0.960$\pm$0.013 & \multicolumn{1}{l|}{} & 0.982$\pm$0.019 & 0.975$\pm$0.027 & 0.978$\pm$0.023 &  \\
\multicolumn{1}{l|}{2HL} & 0.965$\pm$0.016 & 0.961$\pm$0.027 & 0.972$\pm$0.021 & \multicolumn{1}{l|}{0.291} & 0.981$\pm$0.020 & 0.977$\pm$0.027 & 0.978$\pm$0.023 & 1.000 \\ \hline
\multicolumn{1}{l|}{Random-Dense} & 0.959$\pm$0.014 & 0.958$\pm$0.018 & 0.960$\pm$0.016 & \multicolumn{1}{l|}{1.000} & 0.982$\pm$0.019 & 0.975$\pm$0.027 & 0.978$\pm$0.023 & 1.000 \\
\multicolumn{1}{l|}{Random-Sparse} & 0.958$\pm$0.009 & 0.954$\pm$0.021 & 0.958$\pm$0.011 & \multicolumn{1}{l|}{0.829} & 0.990$\pm$0.014 & 0.987$\pm$0.019 & 0.989$\pm$0.015 & 0.397 \\ \hline
\multicolumn{1}{l|}{Bayesian Opt.} & 0.982$\pm$0.005 & 0.983$\pm$0.004 & 0.985$\pm$0.003 & \multicolumn{1}{l|}{{\color[HTML]{FE0000} \textbf{0.003}}} & 0.989$\pm$0.016 & 0.992$\pm$0.011 & 0.989$\pm$0.015 & 0.397
\end{tabular}
\end{table*}


DJINN has consistently higher predictive performance than the two hidden layer model for the regression datasets; the p-values indicate the improvements of DJINN are statistically significant for two of the four datasets. DJINN often achieves slightly higher predictive accuracy for classification tasks, but the improvements over the 2HL model are not statistically significant. 

In general, the performance of DJINN is comparable to existing methods for mapping trees to initialized neural networks for simple datasets, but has higher predictive accuracy for regression tasks. As the complexity of the data increases, it is expected that the deep structure of DJINN will have advantages over the wide, shallow networks, which tend to require more data and time to train \cite{GoodfellowML}.

\subsection{DJINN as a warm-start for neural network training}
Many graph-based models, including decision trees and neural networks, are trained to learn dependency structures in the data. In unsupervised applications, relationships between features are used to find lower-dimensional representations of data \cite{dependency}; in supervised learning, dependency structures relate the input data to output quantities of interest via a series of latent representations formed in the hidden layers of the network \cite{latentdepend}. If an informative structure is initially imposed on the graph, the training process can be accelerated as the imposed relationships act as a warm-start. A common method to warm-start neural networks is the use of a previously trained model to initialize a new model that will be trained on similar, or additional, data. This type of warm-start is often used in transfer learning; it leverages previously-discovered relationships between the inputs, latent representations of the data, and the outputs to accelerate the training process \cite{transfer1}.

The DJINN and 2HL algorithms leverage the dependency structure learned by a decision tree, which has been trained on the data, to warm-start the training of a neural network. By beginning the training process in a state that is primed with dependency information between the input and output data, the tree-based models often converge to a minimum cost in fewer training epochs than randomly initialized networks with the same architecture.

To illustrate the importance of the DJINN weight initialization, the algorithm is compared to other weight initialization schemes. There are two main aspects of DJINN's initial weight topology: the sparsity of the nonzero weights, and where the nonzero weights are placed. To evaluate the importance of the dependency structure imposed by the DJINN weights, the algorithm is compared to neural networks that have no imposed dependency structure: networks with the same architecture, but densely-connected Xavier-initialized weights. To demonstrate that it is not just the sparsity of nonzero weights that is important, but the placement of these weights, DJINN is compared to a network with the same architecture, but with a random, sparse dependency structure imposed on the initial weights. The sparse-random initialization has the same number of nonzero weights per layer that the DJINN initialization utilizes, but with those weights placed randomly within the layer. The initialization guarantees that every neuron has at least one nonzero incoming and outgoing weight; this prevents the initialization from inadvertently changing the architecture by creating neurons that are unable to learn. Like DJINN, the non-zero weights are pulled from the Xavier normal distribution described in section II B.

The middle sections of Tables \ref{table:reg} and \ref{table:class} show the performance of the random-dense and random-sparse initialization schemes for the four regression and classification datasets, respectively. Similar to the comparison between DJINN and the 2HL model, the random initializations are treated as ensemble models: each model contains ten individual neural networks (for DJINN, this corresponds to ten trees, for random initializations this corresponds to ten random seeds used to initialize and place the weights). The prediction from the ensemble is the average of the ten individual predictions. The process of training and evaluating the performance of the random-dense and random-sparse initializations is repeated five times, with the same training and testing datasets used in the comparison between DJINN and the 2HL model. The randomly-initialized networks use the same architectures as DJINN, and are trained with the same hyper-parameters summarized in Table \ref{table:hp}. 

\begin{figure}
\begin{center}
		\includegraphics[width=0.5\textwidth]{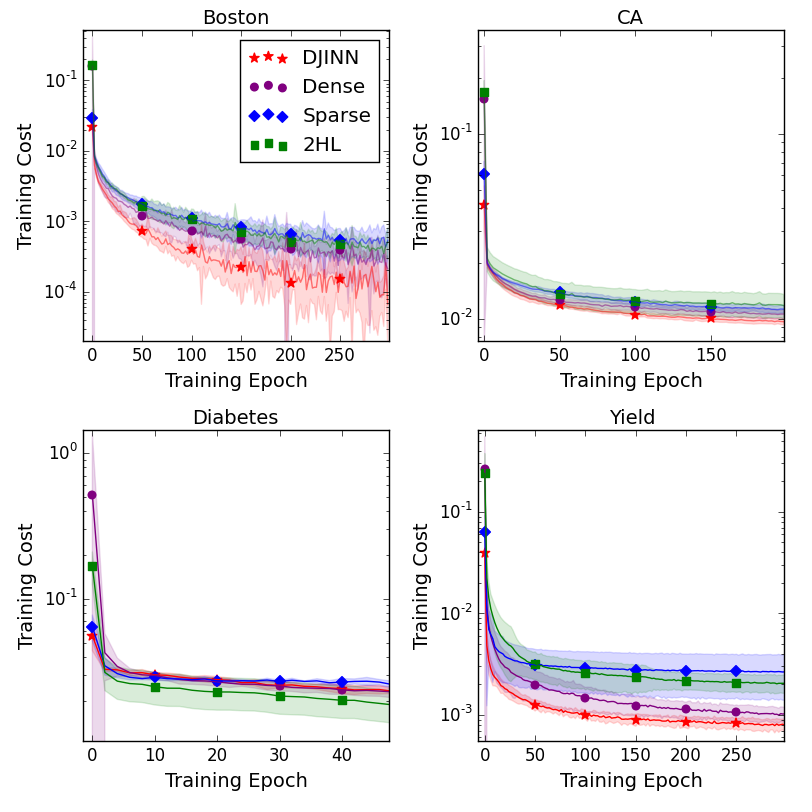}
		\caption{Cost (MSE for data scaled to (0,1)) as a function of training epoch for regression datasets. DJINN weights are observed to start at, and often converges to, a lower cost than the shallow network, or networks with DJINN architecture and other weight initialization techniques. }
		\label{fig:regconv}
 \end{center}
\end{figure}

Figure \ref{fig:regconv} shows the training cost as a function of epoch for the regression tasks; DJINN acts as a warm-start to the training process by consistently starting at a lower cost than other initialization methods. Furthermore, DJINN often converges to the lowest cost, suggesting the network is initialized near a lower local minimum than random initializations can reach in a limited number of training epochs.
The warm-start provided by the decision tree structure leads to higher predictive performance for DJINN in three of the four regression tasks. The improvements of DJINN over the other initialization schemes are statistically significant for the CA housing and yield datasets; the advantages of DJINN are less significant for Boston housing due to the noise in the training cost versus epoch, and the differences in initialization schemes are not significantly different for the diabetes progression data.

While DJINN achieves good predictive accuracy in classification tasks, the advantages of the DJINN weight initialization are less significant. The classification tasks considered are simpler than the regression datasets, thus the performance of various models is less sensitive to the choice of initial weights and hyper-parameter settings. Furthermore, the decision trees are kept shallow due to the size and dimensionality of the datasets; this limits the amount of information mapped from the decision tree into the initialized DJINN model. 

The effects of limiting the depth of the decision tree for datasets with a large number of inputs are illustrated by the digit classification task. Each digit is 64-pixel image that are inputs for the decision tree. The decision tree will split first on the pixel that best separates the digits, however, it is unlikely that a single pixel can provide a significant amount of information about the class to which the image belongs. The decision tree needs to grow deep enough to consider dozens of pixel values before it can accurately classify the image as digit. 
For DJINN, the width of the hidden layers reflects the width of the input layer; with 64 input values, the depth of the neural network must be limited, otherwise there will not be enough data to train the model without a severe risk of over-fitting. Table \ref{table:arch} lists example DJINN architectures for each dataset; indeed, the hidden layers in the digit classification model are wide compared to models with fewer input parameters. 

\begin{figure}
\begin{center}
		\includegraphics[width=0.5\textwidth]{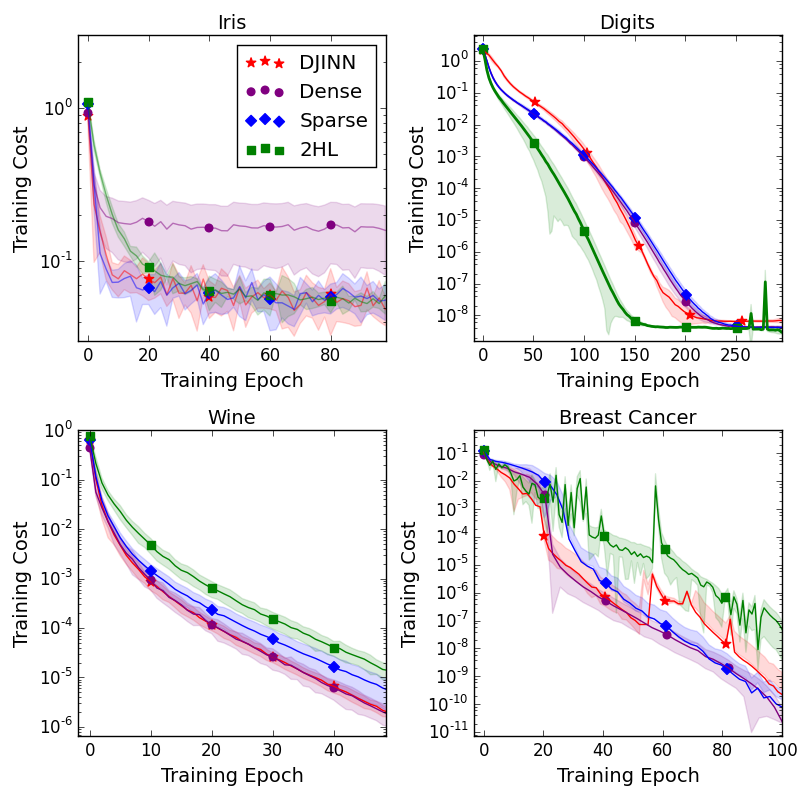}
		\caption{Cost (cross entropy with logits) as a function of training epoch for classification datasets.}
		\label{fig:classconv}
 \end{center}
\end{figure}

With a limited tree depth and a large number of input parameters, the decision paths in the tree are unlikely to contain a significant amount of information to provide a good warm-start for the neural network training procedure. This is illustrated in Fig. \ref{fig:classconv}, where DJINN starts at a cost comparable to the other models for digit classification. In contrast, the iris dataset has four input parameters and three classes; thus the first few splits in the decision tree are able to provide valuable information for separating the classes, and DJINN starts at a slightly lower cost than the other models. 

To handle datasets with a large number of inputs, it would be best to first send the data through convolutional filters or an autoencoder to compress the features into a low-dimensional, meaningful latent space. The latent variables can then be used as inputs to DJINN to build a predictive model. 

To summarize, the benefits of DJINN are most obvious when the trees are sufficiently deep, and the number of nodes in the tree exceeds the number of input parameters; this results in a meaningful dependency structure in the tree that is mapped to initial weights. These conditions are often met for regression tasks, where the warm-start provided by the decision tree allows DJINN to achieve higher predictive performance than non-informative initial weights. Although DJINN does not provide as significant of a warm-start for classification tasks, Table \ref{table:class} shows that DJINN achieves good predictive performance and has the advantage of not requiring the user to hand-tune the architectures for each dataset.

\section{Comparison of DJINN to hyper-parameter optimization}
The utility of DJINN is its ability to be applied as a black box algorithm for efficiently creating accurate deep neural networks for arbitrary datasets. Recently, researchers have started developing a variety of hyper-parameter optimization techniques for designing deep neural networks \cite{automl,reinforce,genetic, clusteringNN}. These algorithms eliminate the need to hand-tune architectures by searching through various combinations of hidden layers and neurons per layer to find the best architecture for a given dataset. Although effective, high-dimensional hyper-parameter searches can be prohibitively expensive. For each proposed architecture, the neural network must be trained to determine the quality of model, and unless the architecture space is restricted to a fixed number of layers or has a constraint on the number of neurons per layer, finding the optimal architecture can require training hundreds of candidate neural networks.

DJINN does not attempt to find the ``optimal'' architecture for a given dataset, it uses an architecture determined by the structure of a decision tree. This architecture, combined with the weight-initialization that leverages the dependency structure from the tree, is observed to produce accurate models for a variety of datasets. 

Although DJINN does not solve the same problem a hyper-parameter optimization method seeks to solve, both methods attempt to improve the usability of neural networks by reducing the number of hyper-parameters that must be specified to train a neural network. It is interesting to see how DJINN, which only requires training a decision tree to propose a suitable architecture, compares to a network designed via architecture optimization. 

The final sections of Tables \ref{table:reg} and \ref{table:class} show the performance of neural networks designed via Bayesian hyper-parameter optimization \cite{bayesopt}. To constrain the search space, the optimizer is restricted to neural networks with the same number of layers used in the DJINN models, and searches for the optimal number of neurons for each hidden layer. Table \ref{table:arch} lists five of the architectures (resulting from the five-fold cross-validation) found via Bayesian optimization, and an example architecture from DJINN for each of the cross-validation steps. The candidate neural networks are trained with the same hyper-parameters summarized in Table \ref{table:hp} and are initialized with Xavier weights. The optimizer is stopped after it has evaluated 100 architectures, and the best model is used to compute the integrated performance metrics. Consistent with the other comparisons, the Bayesian optimizer is run for the five permutations of training/testing datasets to compute cross-validation scores recorded in Tables \ref{table:reg} and \ref{table:class}.

DJINN has a higher predictive performance than the Bayesian optimizer for three of the four regression tasks. The p-values indicate that the improvement of DJINN over the Bayesian optimizer is statistically significant for the CA housing and yield datasets, but the Bayesian optimizer performance is significantly better than DJINN for the diabetes progression data. For classification tasks, the Bayesian optimizer and DJINN perform similarly. Table \ref{table:arch} indicates that the optimization algorithm prefers smaller networks than DJINN for classification tasks; the inclusion of too many degrees of freedom in DJINN could explain its lower performance for the breast cancer and wine classification tasks, consistent with previous discussions.

Computational efficiency is important to consider when employing hyper-parameter optimization procedures. For the examples presented above, the hyper-parameter optimization algorithm evaluates 100 neural networks; this requires approximately 10x the training time of DJINN when the ten network ensemble is trained serially, or 100x the training time of DJINN if the ten networks are trained in parallel. 
For the moderate-sized datasets, hyper-parameter optimization is feasible. However, for high volume, high dimensional datasets, hyper-parameter searches become prohibitively expensive. DJINN remains comparatively inexpensive as the complexity of the data increases, requiring only the construction of a small ensemble of decision trees, which are often trained in seconds, to determine an appropriate architecture and weight initialization. Subsequent training of the individual neural networks in a DJINN ensemble can then be carried out in parallel, offering significant advantages over sequential hyper-parameter optimization methods. 


\begin{table}[]
\centering
\caption{Hidden layer widths from DJINN and a Bayesian optimizer for each dataset. The width of the input layer reflects the number of features in each dataset. The output layer has a single neuron for regression tasks, and one neuron per class for classification tasks.}
\label{table:arch}
\resizebox{0.5\textwidth}{!}{
\begin{tabular}{ll|l}
Dataset & DJINN & Bayesian Opt. \\ \hline
Boston & \begin{tabular}[c]{@{}l@{}}(15,17,20,18), (15,17,20,18), \\ (13,15,22,27), (15,18,24,18), \\ (14,18,22,26)\end{tabular} & \begin{tabular}[c]{@{}l@{}}(7,10,9,14), (7,10,8,14), \\ (13,14,10,7), (7,11,7,15), \\ (12,9,8,11)\end{tabular} \\ \hline
CA Housing & \begin{tabular}[c]{@{}l@{}}(10,12,19,23), (10,11,16,25), \\ (10,11,19,25), (10,11,17,26), \\ (10,14,19,21)\end{tabular} & \begin{tabular}[c]{@{}l@{}}(12,20,5,6), (4,17,14,16), \\ (5,13,16,5), (12,14,8,5), \\ (20,5,20,8)\end{tabular} \\ \hline
Diabetes & \begin{tabular}[c]{@{}l@{}}(12,15,19,28), (12,14,19,28), \\ (11,15,19,28), (12,15,22,23), \\ (11,14,20,22)\end{tabular} & \begin{tabular}[c]{@{}l@{}}(4,1,4,7,9), (9,9,19,7), \\ (18,9,10,4), (11,16,13,18), \\ (13,15,16,18)\end{tabular} \\ \hline
Yield & \begin{tabular}[c]{@{}l@{}}(11,15,18,23), (11,14,19,21), \\ (10,15,22,25), (10,15,21,25), \\ (10,13,21,23)\end{tabular} & \begin{tabular}[c]{@{}l@{}}(5,5,30,30), (20,12,15,5), \\ (22,10,23,19), (5,5,16,5), \\ (14,18,11,14)\end{tabular} \\ \hline
Iris & (5,4), (5,7), (5,5), (5,7), (4,5) & \begin{tabular}[c]{@{}l@{}}(14,6), (11,7), (13,7), \\ (10,8), (11,4)\end{tabular} \\ \hline
Digits & \begin{tabular}[c]{@{}l@{}}(65,48), (64,33), (63,33), \\ (63,66), (64,23)\end{tabular} & \begin{tabular}[c]{@{}l@{}}(17,32), (13,31), (5,49), \\ (5,28), (13,31)\end{tabular} \\ \hline
Wine & \begin{tabular}[c]{@{}l@{}}(14,15), (15,11), (15,9), \\ (15,11), (13,9)\end{tabular} & \begin{tabular}[c]{@{}l@{}}(4,11), (12,14), (7,12), \\ (10,12), (12,5)\end{tabular} \\ \hline
Breast Cancer & \begin{tabular}[c]{@{}l@{}}(32,33,19), (32,30,18), \\ (32,33,23), (32,30,24), \\ (30,31,19)\end{tabular} & \begin{tabular}[c]{@{}l@{}}(5,3,6), (6,5,5), (2,5,6), \\ (6,4,2), (5,6,4)\end{tabular}
\end{tabular}}
\end{table}

Overall, there is compelling empirical evidence to suggest DJINN is a robust black box algorithm for creating accurate neural networks for a wide variety of datasets. The advantages of DJINN are most evident in complex regression problems, where the choice of architecture and initialization can greatly impact the predictive performance of the model. For simple classification problems, the performance of DJINN is comparable to other network design and weight initialization techniques. Although DJINN is not attempting to find an optimal architecture, when compared against hyper-parameter optimization for designing neural networks, DJINN displays competitive performance while requiring significantly lower computational costs. DJINN successfully combines the usability of decision tree models with the flexibility of deep neural networks to produce accurate predictive models for a variety of problems.


\section{Conclusions}
The flexibility and powerful predictive capabilities of neural networks are combined with user-friendly decision tree models to create scalable and accurate ``deep jointly-informed neural networks'' (DJINN). The DJINN algorithm maps an ensemble of decision trees trained on a dataset into an ensemble of initialized neural networks that are subsequently trained via back-propagation. 
The information mapped from the decision trees into initial weights provides a warm-start to the neural network training process; thus DJINN is often observed to start at, and converge to, a lower cost than other neural network initialization methods. 

DJINN reduces the number of user-specified hyper-parameters needed to create a deep neural network by using the decision tree structure to determine the network architecture. When compared to hyper-parameter optimization methods for selecting an appropriate architecture, DJINN displays competitive performance at a fraction of the computational cost, demonstrating that an optimal architecture is not necessary if the weight initialization is sufficiently informative. 

Although formulated for fully-connected feed-forward neural networks, DJINN could also be applied to networks that use feed-forward neural networks as part of a more complex system. For example, DJINN could be used for image analysis tasks after convolutional layers extract the important features. 
By combining the ease of use of decision trees with the predictive power of deep neural networks, DJINN is an attractive method for easily creating surrogate models of complex systems.

\section*{Acknowledgment}
The authors would like to thank Jayaraman Thiagarajan, Ryan Nora, Brian Spears, John Field, Jim Gaffney, Michael Kruse, Scott Brandon, Paul Springer, and Jim Brase for fruitful discussions. This work was performed under the auspices of the U.S. Department of Energy by Lawrence Livermore National Laboratory under Contract DE-AC52-07NA27344. Document released as LLNL-JRNL-732588.
\bibliography{djinn}
\bibliographystyle{IEEEtran}

\end{document}